%% file: main.tex
\documentclass[sigconf]{acmart}

\usepackage{color,colortbl}
\usepackage{multirow,paralist,xspace}
\AtBeginDocument{%
  \providecommand\BibTeX{{%
    \normalfont B\kern-0.5em{\scshape i\kern-0.25em b}\kern-0.8em\TeX}}}

\copyrightyear{2021} 
\acmYear{2021} 
\setcopyright{acmcopyright}\acmConference[KDD '21]{Proceedings of the 27th ACM SIGKDD Conference on Knowledge Discovery and Data Mining}{August 14--18, 2021}{Virtual Event, Singapore}
\acmBooktitle{Proceedings of the 27th ACM SIGKDD Conference on Knowledge Discovery and Data Mining (KDD '21), August 14--18, 2021, Virtual Event, Singapore}
\acmPrice{15.00}
\acmDOI{10.1145/3447548.3467308}
\acmISBN{978-1-4503-8332-5/21/08}

\begin{document}

\title{Enhancing Taxonomy Completion with Concept Generation via Fusing Relational Representations}

\author{Qingkai Zeng, Jinfeng Lin, Wenhao Yu, Jane Cleland-Huang, Meng Jiang}
\affiliation{%
  \institution{{Department of Computer Science and Engineering, University of Notre Dame, Notre Dame \state{IN} \country{USA}}\\}
}
\email{{qzeng, jlin6, wyu1, janehuang, mjiang2}@nd.edu, }


\newcommand*{\Scale}[2][4]{\scalebox{#1}{$#2$}}
\newcommand{\edge}[2]{\text{⟨}#1,~#2\text{⟩}}
\newcommand{\method}{\textsc{GenTaxo}\xspace}
\newcommand{\methodplus}{\textsc{GenTaxo++}\xspace}
\newcommand\todo[1]{\textcolor{red}{#1}}
\newcommand\placeholder[1]{\textcolor{blue}{#1}}
\newcommand\wyu[1]{\textcolor{red}{ wenhao:#1}}
\newcommand\jch[1]{\textcolor{blue}{ jane:#1}}
\newcommand\blfootnote[1]{%
  \begingroup
  \renewcommand\thefootnote{}\footnote{#1}%
  \addtocounter{footnote}{-1}%
  \endgroup
}

\begin{abstract}
  A clear and well-documented \LaTeX\ document is presented as an
  article formatted for publication by ACM in a conference proceedings
  or journal publication. Based on the ``acmart'' document class, this
  article presents and explains many of the common variations, as well
  as many of the formatting elements an author may use in the
  preparation of the documentation of their work.
\end{abstract}




\keywords{Taxonomy Completion, Concept Generation}

\begin{abstract}
\input{0Abstract}
\end{abstract}

\maketitle

\blfootnote{$\S$ Code and data is available at \url{https://github.com/QingkaiZeng/GenTaxo}.}

% \section{Introduction}
\vspace{-0.15in}
\input{1Introduction}
\input{3Problem}
\input{4Framework}
\input{5ExperimentDesign}
\input{6ExperimentResults}

\input{2RelatedWork}
% \vspace{-0.05in}
\input{7Conclusions}

\section*{Acknowledgment}
This research was supported by National Science Foundation award CCF-1901059.

\balance
\bibliographystyle{ACM-Reference-Format}
\bibliography{main}

\clearpage
\balance
\appendix
\section{Appendix}
\input{aAppendix}

\end{document}

%% file: 0Abstract.tex
Automatic construction of a taxonomy supports many applications in e-commerce, web search, and question answering. Existing taxonomy expansion or completion methods assume that new concepts have been accurately extracted and their embedding vectors learned from the text corpus. 
However, one critical and fundamental challenge in fixing the incompleteness of taxonomies is  the incompleteness of the extracted concepts, especially for those whose names have multiple words and consequently low frequency in the corpus. 
To resolve the limitations of  extraction-based methods, we propose \method to enhance taxonomy completion by identifying positions in existing taxonomies that need new concepts and then generating appropriate concept names. Instead of relying on the corpus for concept embeddings, \method learns the contextual embeddings from their surrounding graph-based and language-based relational information, and leverages the corpus for pre-training a concept name generator. Experimental results demonstrate that \method improves the completeness of taxonomies over existing methods.

%% file: 1Introduction.tex
\section{Introduction}
\label{sec:introduction}

Taxonomies have been widely used to enhance the performance of many applications such as question answering \cite{yang2017efficiently,yu-etal-2021-technical} and personalized recommendation \cite{huang2019taxonomy}.
With the influx of new content in evolving applications, it is necessary to curate these taxonomies to include emergent concepts; however, manual curation is labor-intensive and time-consuming.
To this end, many recent studies aim to automatically expand or complete an existing taxonomy.
For example, given a new concept, Shen \emph{et al.} measured the likelihood  of each existing concept in the taxonomy being its hypernym and then added it as a new leaf node \cite{shen2020taxoexpan}. Manzoor \emph{et al.} extended the measurement to be taxonomic relatedness with implicit relational semantics \cite{manzoor2020expanding}. Zhang \emph{et al.} predicted the position of the new concept considering hypernyms and hyponyms \cite{zhang2021taxonomy}. In all of these cases, the distance between concepts was measured using their embeddings learned from some text corpus, with the underlying assumption that new concepts could be extracted accurately and found frequently in the corpus.

\begin{figure}[t]
    \centering
    \includegraphics[width=0.9\linewidth]{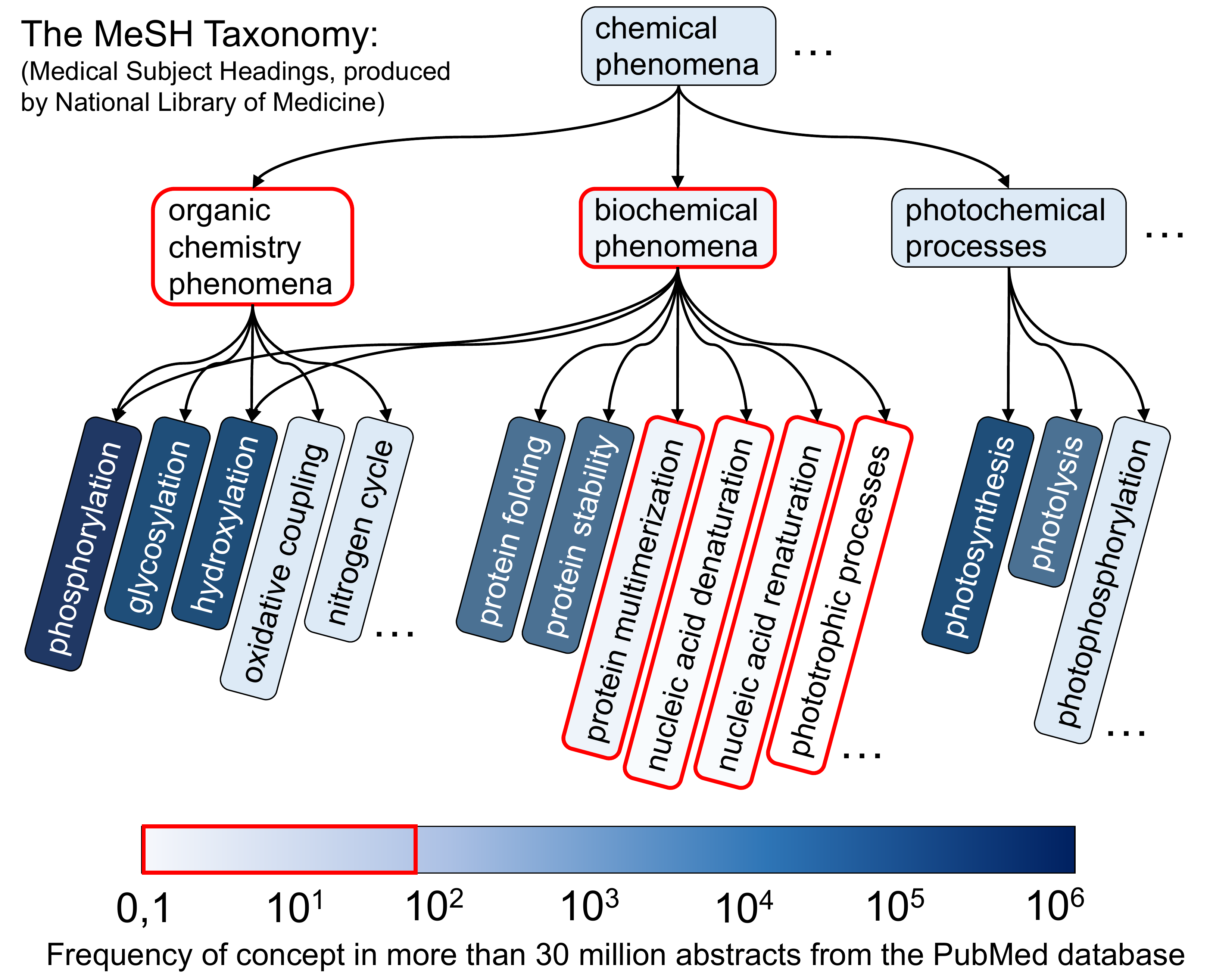}
    \caption{A great number of concepts that are desired to be in the taxonomies (e.g., MeSH) are very rare in even large-scale text corpus (e.g., 30 million PubMed abstracts). So it is hard to extract these concepts or learn their embedding vectors. Concept names need to be generated rather than extracted to fix the incompleteness of taxonomies.}
    \label{fig:motivation}
    \vspace{-0.2in}
\end{figure}

We argue that such an assumption is inappropriate in real-world taxonomies based on the frequency of concepts in Medical Subject Headings (MeSH), a widely-used taxonomy of approximately 30,000 terms that is updated annually and manually, in a large-scale public text corpus of 30 million paper abstracts (about 6 billion tokens) from the PubMed database.
We observe that many concepts that have multiple words appear fewer than 100 times in the corpus (as depicted by the red outlined nodes in Figure~\ref{fig:motivation}) and around half of the terms cannot be found in the corpus (see Table~\ref{tab:datasets} in Section~\ref{sec:5ExperimentDesign}). Concept extraction tools \cite{zeng2020tri} often fail to find them at the top of a list of over half a million concept candidates; and there is insufficient data to learn their embedding vectors.
The incompleteness of concepts is a critical challenge in taxonomy completion, and has not yet been properly studied.

\begin{figure}[t]
    \centering
    \includegraphics[width=0.95\linewidth]{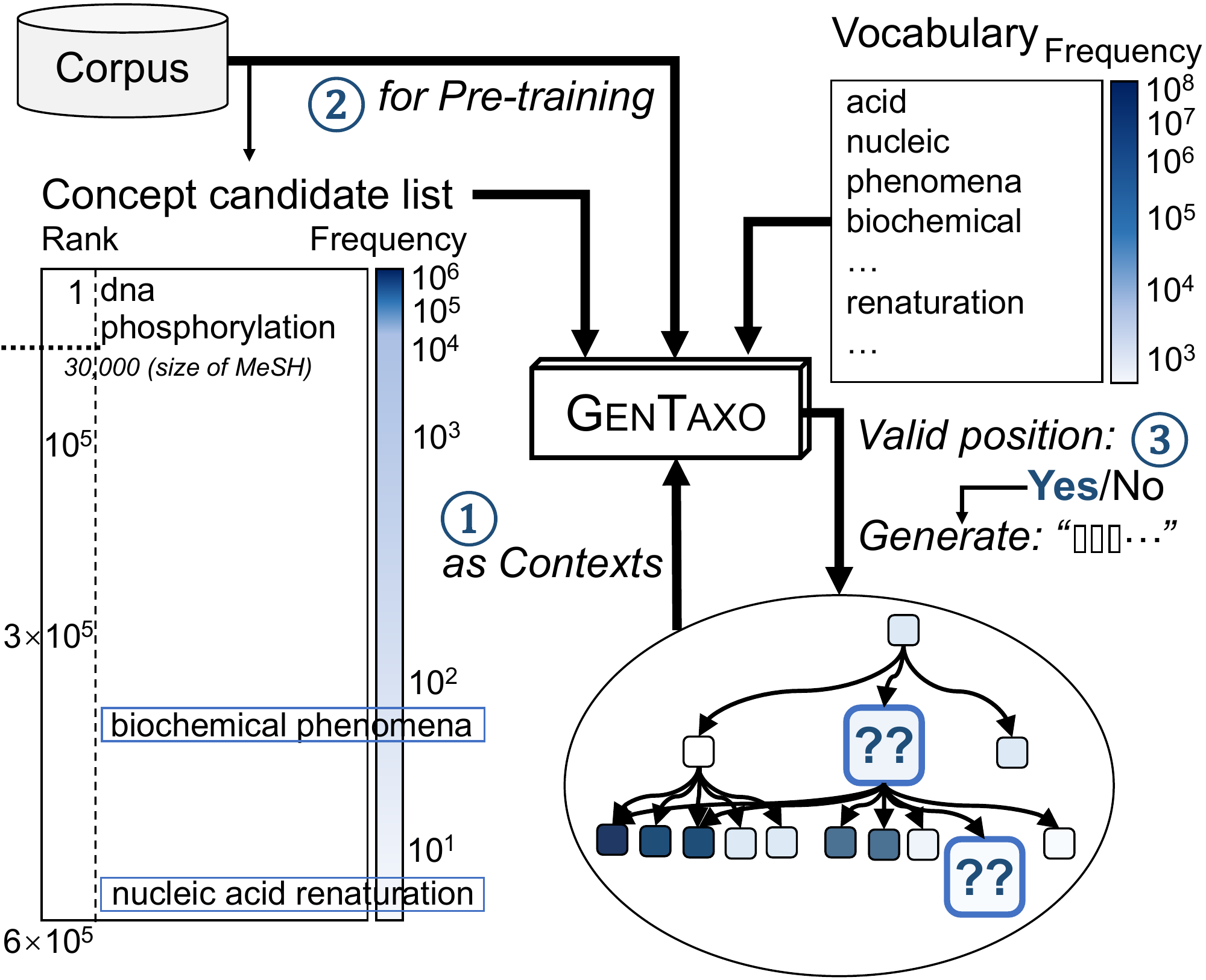}
    \vspace{-0.15in}
    \caption{Existing methods extracted a concept candidate list from the corpus and assumed that the top concepts (high frequency or quality) could be added into existing taxonomies. However, frequent concepts may not be qualified and many qualified concepts are very rare. We find that though the concept names are not frequent, their tokens are. Our idea is to generate concept names at candidate positions with relational contexts to complete taxonomies. Our solution \method is pre-trained on the corpus and learns to generate the names token by token on the existing taxonomy.}
    \label{fig:idea}
    \vspace{-0.2in}
\end{figure}

Despite the low frequency of many multi-gram concepts in a text corpus, the frequency of individual words is naturally much higher. Inspired by recent advances in text generation \cite{meng2017deep,yu2020survey}, we propose a new task, ``taxonomy generation'', that identifies whether a new concept fits in a candidate position within an existing taxonomy, and if it does fit, generates the concept name token by token.

The key challenge lies in the lack of information for accurately generating the names of new concepts when their full names do not (frequently) appear in the text corpus. To address this challenge, our framework for enhancing taxonomy completion, called \method, has the following novel design features (see Figure~\ref{fig:idea} and~\ref{fig:framework}):

First, \method has an encoder-decoder scheme that \emph{learns to generate} any concept in the existing taxonomy in a self-supervised manner. Suppose an existing concept $v$ is masked. Two types of encoders are leveraged to fuse both sentence-based and graph-based representations of the masked position learned from the relational contexts. One is a sequence encoder that learns the last hidden states of a group of sentences that describe the relations such as ``$v_p$ is a class of'' and ``$v_c$ is a subclass of'', when $v_p$ and $v_c$ are parent and child concepts of the position, respectively. The other is a graph encoder that aggregates information from two-hop neighborhoods in the top-down subgraph (with $v$ at the bottom) and bottom-up subgraph (with $v$ at the top). The fused representations are fed into a GRU-based decoder to generate $v$'s name token by token with a special token [EOS] at the end. With the context fusion, the decoding process can be considered as the completion of a group of sentences (e.g., ``$v_p$ is a class of'' and ``$v_c$ is a subclass of'') with the same ending term while simultaneously creating a hyponym/hypernym node in the top-down/bottom-up subgraphs. 

Second, \method is pre-trained on a large-scale corpus to predict tokens in concept names. The pre-training task is very similar to the popular Mask Token Prediction task \cite{devlin2019bert} except that the masked token must be a token of a concept that appears in the existing taxonomy and is found in a sentence of the corpus.

Third, \method performs the task of candidate position classification simultaneously with concept name generation. It has a binary classifier that uses the final state of the generated concept name to predict whether the concept is needed at the position. We adopt negative sampling to create ``invalid'' candidate positions in the existing taxonomy. The classifier is attached after the name generation (not before it) because the quality of the generated name indicates the need for a new concept at the position.

Furthermore, we develop \methodplus to enhance extraction-based methods, when a set of new concepts, though incomplete, needs to be added to existing taxonomies (as described in existing studies). In \methodplus, we apply \method to generate concept names in order to expand the set of new concepts. We then use the extraction-based methods with the expanded set to improve the concept/taxonomy completeness.

The main contributions of this work are summarized as follows:
\begin{itemize}
\item We propose a new taxonomy completion task that identifies valid positions to add new concepts in existing taxonomies and generates their names token by token.
\item We design a novel framework \method that has three novel designs: (1) an encoder-decoder scheme that fuses sentence-based and graph-based relational representations and learns to generate concept names; (2) a pre-training process for concept token prediction; and (3) a binary classifier to find valid candidate positions. Furthermore, \methodplus is developed to enhance existing extraction-based methods with the generated concepts, when a set of new concepts are available.
\item Experiments on \emph{six} real-world taxonomy data sets demonstrate that (1) concept generation can significantly improve the recall and completeness of taxonomies; (2) even some concepts that do not appear in the corpus can be accurately generated token by token at valid positions; and (3) fusing two types of relational representations is effective. 
\end{itemize}

%% file: 3Problem.tex
\section{Problem Definition}

\begin{figure*}[t]
    \centering
    \includegraphics[width=0.92\linewidth]{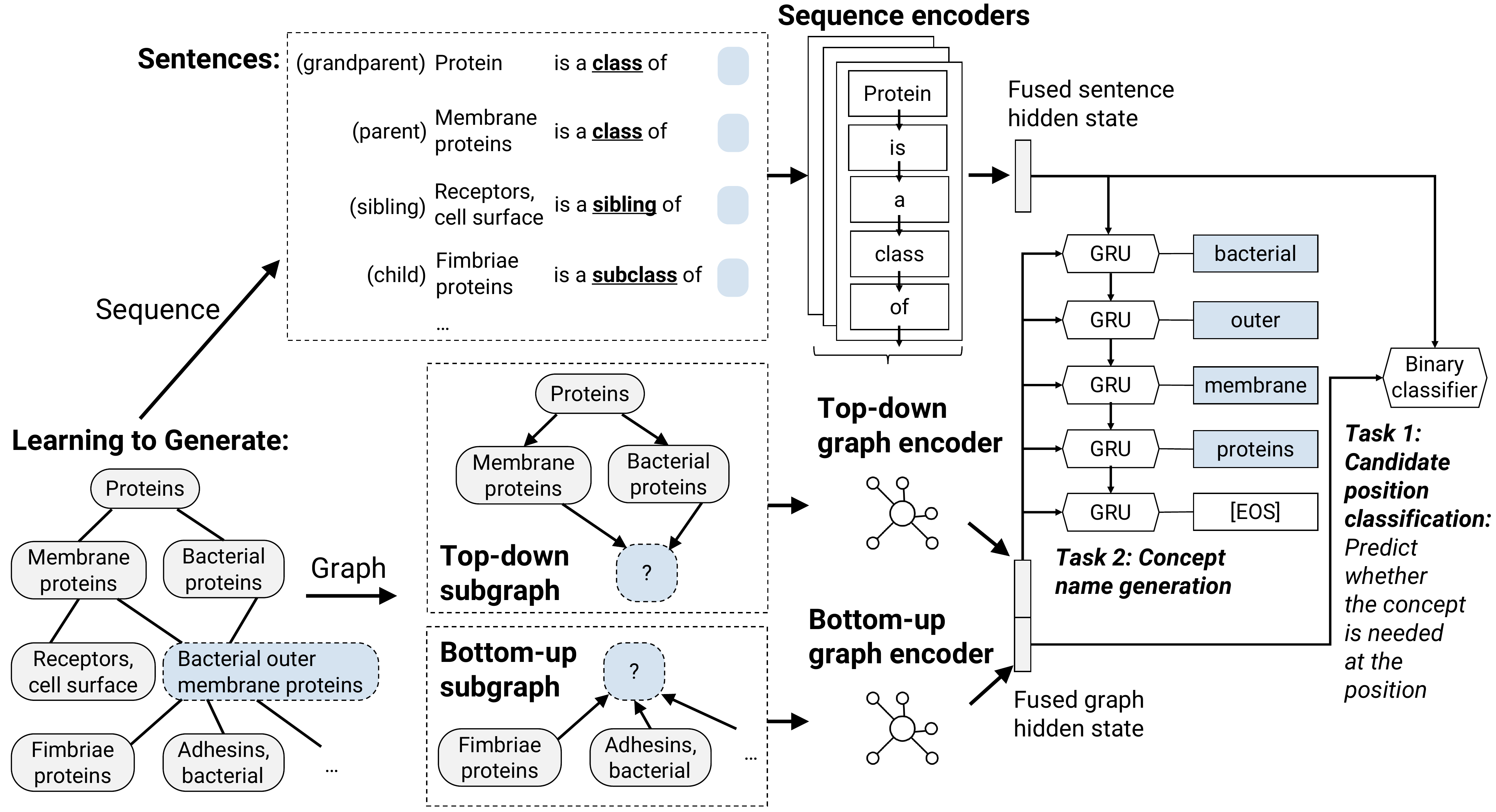}
    \vspace{-0.1in}
    \caption{Architecture of \method.
    Suppose the concept ``bacterial outer membrane proteins'' is masked. Given the masked position in the taxonomy, the model learns to generate the name of the concept.
    It has two types of encoders to represent the relational contexts of the position.
    Sequence encoders learn a group of sentences that describe the relations with the masked concept.
    Graph encoders learn the masked concept's embedding by aggregating information from its top-down and bottom-up subgraphs.
    The fused sentence- and graph-based hidden states are used to predict whether the masked position is valid (Task 1).
    They are fed into GRU decoders to generate the concept's name (Task 2).}
    \label{fig:framework}
\end{figure*}

Traditionally, the input of the taxonomy completion task includes two parts \cite{manzoor2020expanding,shen2020taxoexpan,yu2020steam,zhang2021taxonomy}: (1) an existing taxonomy $\mathcal{T}_0 = (\mathcal{V}_0, \mathcal{E}_0)$ and (2) a set of new concepts $\mathcal{C}$ that have been either manually given or accurately extracted from text corpus $\mathcal{D}$.  The overall goal is completing the existing taxonomy $\mathcal{T}_0$ into a larger one $\mathcal{T} = (\mathcal{V}_0 \cup \mathcal{C}, \mathcal{E}^{'}$).

We argue that existing technologies, which defined the task as described above, could fail when concepts cannot be extracted due to their low frequency, or cannot be found in the corpus. We aim to mitigate this problem by \emph{generating} the absent concepts to achieve better taxonomy completion performance. Suppose $v = (t_1, t_2, \dots, t_{T_v})$, where $t_i \in \mathcal{W}$ is the $i$-th token in the name of concept $v$, $T_v$ is the number of $v$'s tokens (length of $v$), $\mathcal{W}$ is the token vocabulary which includes words and punctuation marks (e.g., comma).

\begin{definition}[Taxonomy]
We follow the definition of taxonomy in \cite{zhang2021taxonomy}. A taxonomy $\mathcal{T} = (\mathcal{V}, \mathcal{E})$ is a directed acyclic graph where each node $v \in  \mathcal{V}$ represents a concept (i.e., a word or a phrase) and each directed edge $\edge{u}{v} \in \mathcal{E}$ implies a relation between a parent-child pair such as ``is a type of'' or ``is a class of''. We expect the taxonomy to follow a hierarchical structure where concept $u$ is the most specific concept that is more general than concept $v$. Note that a taxonomy node may have multiple parents.
\end{definition}

In most taxonomies, the parent-child relation can be specified as a hypernymy relation between concept $u$ and $v$, where $u$ is the hypernym (parent) and $v$ is the hyponym (child). We use the terms ``hypernym'' and ``parent'', ``hyponym'' and ``child'' interchangeably throughout the paper.

\begin{definition}[Candidate Position]
Given two sets of concepts $\mathcal{V}_p, \mathcal{V}_c \subset \mathcal{V}_0$, if $v_c$ is one of the descendants of $v_p$ in the existing taxonomy for any pair of concepts $v_p \in \mathcal{V}_p$ and $v_c \in \mathcal{V}_c$, then a candidate position acts as a placeholder for a new concept $v$. It becomes a valid position when $v$ satisfies (1) $v_p$ is a parent of $v$ and (2) $v_c$ is a child of $v$. When it is a valid position, we add edges $\edge{v_p}{v}$ and $\edge{v}{v_c}$ and delete redundant edges to update $\mathcal{E}^{'}$.
\end{definition}

We reduce the task of generating concept names for taxonomy completion as the problem below: Given text corpus $\mathcal{D}$ and a candidate position on an existing taxonomy $\mathcal{T}_0$, predict whether the position is valid, and if yes, generate the name of the concept $v$ from the token vocabulary $\mathcal{W}$ (extracted from $\mathcal{D}$) to fill in the position.

%% file: 4Framework.tex
\section{\method: Generate Concept Names}

\paragraph{Overall architecture} Figure~\ref{fig:framework} presents the architecture of \method. The goal is to learn from an existing taxonomy to identify valid candidate positions and generate concept names at those positions. Given a taxonomy, it determines \emph{valid} candidate positions by masking an existing concept in the taxonomy; it also determines \emph{invalid} candidate positions using negative sampling strategies which will be discussed in Section~\ref{sec:decoder}.

Here we focus on the valid positions and masked concepts. As shown in the left bottom side of Figure~\ref{fig:framework}, suppose the term ``bacterial outer membrane proteins'' is masked. \method adopts an encoder-decoder scheme in which the encoders represent the taxonomic relations in forms of sentences and subgraphs (see the middle part of the figure) and the decoders perform two tasks to achieve the goal (see the right-hand side of the figure).

\subsection{Encoders: Representing Taxonomic Relations}

Here we introduce how to represent the taxonomic relations into sentences and subgraphs, and use sequence and graph encoders to generate and fuse the relational representations.

\subsubsection{Representing Taxonomic Relations as Sentences}

Given a taxonomy $\mathcal{T}_0 = (\mathcal{V}_0, \mathcal{E}_0)$ and a candidate position for masked concept $v$, we focus on three types of taxonomic relations between the masked concept $v$ and some concepts $u \in \mathcal{V}_0$:

\begin{enumerate}
\item \emph{Parent or ancestor.} Suppose there is a sequence of nodes $(v_1, v_2, \dots, v_l)$ where $v_1 = u$, $v_l = v$, and $\edge{v_i}{v_{i+1}} \in \mathcal{E}_0$ for any $i = 1, \dots, l-1$. In other words, there is a path from $u$ to $v$ in $\mathcal{T}_0$. When $l=1$, $u$ is a parent of $v$; when $l \geq 2$, $u$ is an ancestor. We denote $u = (t_1, t_2, \dots, t_{T_u})$, where $T_u$ is the length of concept $u$ and $\{t_i\}^{T_u}_{i=1}$ are its tokens. Then we create a sentence with the template below:
\begin{quote}
    ``$t_1$ $t_2$ $\dots$ $t_{T_u}$ is a class of [MASK]''
\end{quote}
And we denote the set of parent or ancestor nodes of $v$ (i.e., all possible $u$ as described above) by $\mathcal{V}_p (v)$.
\item \emph{Child or descendant.} Similar as above expect $v_1 = v$ and $v_l = u$, we have a path from $v$ to $u$. When $l=1$, $u$ is a child of $v$; when $l \geq 2$, $u$ is a descendant. Then we create a sentence:
\begin{quote}
    ``$t_1$ $t_2$ $\dots$ $t_{T_u}$ is a subclass of [MASK]''
\end{quote}
We denote the set of child or descendant nodes of $v$ by $\mathcal{V}_c (v)$.
\item \emph{Sibling.} If there is a concept $p \in \mathcal{V}_0$ and two edges $\edge{p}{v}$ and $\edge{p}{u}$, then $u$ is a sibling node of $v$. We create a sentence:
\begin{quote}
    ``$t_1$ $t_2$ $\dots$ $t_{T_u}$ is a sibling of [MASK]''
\end{quote}
\end{enumerate}

Given the candidate position for concept $v$, we can collect a set of sentences from its relational contexts, denoted by $\mathcal{S}(v)$.
% Examples can be found in the left top dashed box of Figure~\ref{fig:framework}.
% \paragraph{Sentence encoders:}
We apply neural network models that capture sequential patterns in the sentences and create the hidden states at the masked position. The hidden states are vectors that encode the related concept node $u$ and the relational phrase to indicate the masked concept $v$. The hidden states support the tasks of concept name generation and candidate position classification. We provide \emph{three} options of sentence encoders:
% BiGRU, Transformer \cite{vaswani2017attention}, and SciBERT \cite{beltagy2019scibert}. Their descriptions can be found in Appendix.
BiGRU, Transformer, and SciBERT \cite{beltagy2019scibert}.
% Their descriptions can be found in Appendix.

\subsubsection{Representing Taxonomic Relations as Subgraphs}

The relations between the masked concept $v$ and its surrounding nodes can be represented as two types of subgraphs:
% , because of the hierarchical structure of $\mathcal{T}_0$ (see the middle bottom part of Figure~\ref{fig:framework}):

\begin{enumerate}
\item Top-down subgraph: It consists of all parent and ancestor nodes of $v$, denoted by $G_{\text{down}}(v) = \{\mathcal{V}_{\text{down}}(v), \mathcal{E}_{\text{down}}(v)\}$, where $\mathcal{V}_{\text{down}}(v) = \mathcal{V}_p (v)$ and $\mathcal{E}_{\text{down}}(v) = \{\edge{v_i}{v_j} \in \mathcal{E} \mid v_i, v_j \in \mathcal{V}_{\text{down}}(v)\}$.
The role of $v$ is the very specific concept of any other concepts in this subgraph. So, the vector representations of this masked position should be aggregated in a top-down direction. The aggregation indicates the relationship of being from more general to more specific.
\item Bottom-up subgraph: Similarly, it consists of all child and descendant nodes of $v$, denoted by $G_{\text{up}}(v) = \{\mathcal{V}_{\text{up}}(v), \mathcal{E}_{\text{up}}(v)\}$, where $\mathcal{V}_{\text{up}}(v) = \mathcal{V}_c (v)$ and $\mathcal{E}_{\text{up}}(v) = \{\edge{v_i}{v_j} \in \mathcal{E} \mid v_i, v_j \in \mathcal{V}_{\text{up}}(v)\}$.
% The role of $v$ is the very general concept of any other concepts in this subgraph. So,
The representations of this masked position should be aggregated in a bottom-up direction.
The aggregation indicates the relationship of being from specific to general.
\end{enumerate}

\paragraph{Graph encoders:} We adopt two independent graph neural networks (GNNs) to encode the relational contexts in $G_{\text{down}}(v)$ and $G_{\text{up}}(v)$ separately. 
% GNNs are multi-layered neural networks dedicated for graph encoding. They collect contextual information for generating the representations of a target node by aggregating those of its adjacent nodes.Next, we describe the GNN algorithm we use for a subgraph $G$, no matter it is $G_{\text{down}}(v)$ or $G_{\text{up}}(v)$.
Given a subgraph $G$, GNN learns the graph-based hidden state of \emph{every node} on the final layer through the graph structure, while we will use that of the node $v$ for next steps.

$v$ was specifically denoted for the masked concept node. In this paragraph, we denote any node on $G$ by $v$ for convenience. We initialize the vector representations of $v$ randomly, denoted by $\mathbf{v}^{(0)}$. Then, the $K$-layered GNN runs the following functions to generate the representations of $v$ on the $k$-th layer ($k=1, \dots, K$):
\begin{eqnarray}
  \mathbf{a}^{(k-1)} & = & \textsc{Aggregate}_k(\big{\{} (\mathbf{u}^{(k-1)}): \forall u \in \mathcal{N}(v) \cup G \big{\}}), \nonumber \\
  \mathbf{v}^{(k)} & = & \textsc{Combine}_k (\mathbf{v}^{(k-1)}, \nonumber \mathbf{a}^{(k-1)}),
\end{eqnarray}
where $\mathcal{N}(v)$ is the set of neighbors of $v$ in $G$.

There are a variety of choices for $\textsc{Aggregate}_k(\cdot)$ and $\textsc{Combine}_k(\cdot)$.
For example, $\textsc{Aggregate}_k(\cdot)$ can be mean pooling, max pooling, graph attention, and concatenation~\cite{koncel2019text,hamilton2017inductive}.
One popular choice for $\textsc{Combine}_k(\cdot)$ is graph convolution: $\mathbf{v}^{(k)} = \sigma \left(\mathbf{W}^{(k)} (\mathbf{v}^{(k-1)} \oplus \mathbf{a}^{(k-1)}) \right)$,
where $\mathbf{W}^{(k)}$ is the weight matrix for linear transformation on the $k$-th layer and $\sigma(\cdot)$ is a nonlinear function.

For the masked concept $v$, we finally return the graph-based hidden state $\mathbf{v}_G$ at K-th layer, i.e., $\mathbf{v}_G = \mathbf{v}^{(K)}$.

\subsubsection{Representations Fusion}

As aforementioned, we have a set of sentence-based hidden states $\{\mathbf{h}(s)\}_{s \in \mathcal{S}(v)}$ from the sentence encoder; also, we have two graph-based hidden states $\mathbf{v}_{G_{\text{down}}}$ and $\mathbf{v}_{G_{\text{up}}}$ from the graph encoder. In this section, we present how to fuse these relational representations for decoding concept names.

\paragraph{Fusing sentence-based hidden states:} We use the self-attention mechanism to fuse the hidden states with a weight matrix $\mathbf{W}_{\text{seq}}$:
\begin{equation}
    \mathbf{h}_{\text{seq}} = \sum_{s \in \mathcal{S}(v)} w(s) \cdot \mathbf{h}(s), ~\text{where}~ w(s) = \frac{\exp( \sigma(\mathbf{W}_{\text{seq}}\mathbf{h}(s)))}{\sum\limits_{s' \in \mathcal{S}(v)} \exp(\sigma(\mathbf{W}_{\text{seq}}\mathbf{h}(s')))}. \nonumber
\end{equation}

\paragraph{Fusing graph-based hidden states:} We adopt a learnable weighted sum to fuse $\mathbf{v}_{G_{\text{down}}}$ and $\mathbf{v}_{G_{\text{up}}}$ with weight matrices $\mathbf{W}_{\text{down}}$ and $\mathbf{W}_{\text{up}}$:
\begin{equation}
    \mathbf{v}_{\text{graph}} = \beta \cdot \mathbf{v}_{G_{\text{down}}} + (1-\beta) \cdot \mathbf{v}_{G_{\text{up}}}, \nonumber
\end{equation}
where $\beta = \frac{\exp(\sigma(\mathbf{W}_{\text{down}} \cdot \mathbf{v}_{G_{\text{down}}}))}{\exp(\sigma(\mathbf{W}_{\text{down}} \cdot \mathbf{v}_{G_{\text{down}}})) + \exp(\sigma(\mathbf{W}_{\text{up}} \cdot \mathbf{v}_{G_{\text{up}}}))}$.

\paragraph{Fusing the fused sentence- and graph-based hidden states:} Given a masked concept $v$, there are a variety of strategies to fuse $\mathbf{h}_{\text{seq}}$ and $\mathbf{v}_{\text{graph}}$: $\mathbf{v}_{\text{fuse}} = \textsc{Fuse}(\mathbf{h}_{\text{seq}}, \mathbf{v}_{\text{graph}})$, such as mean pooling, max pooling, attention, and concatenation. Take concatenation as an example: $\textsc{Fuse}(\mathbf{a}, \mathbf{b}) = \mathbf{a} \oplus \mathbf{b}$.
% We will investigate these fusing strategies in the ablation study at the experiments section.

\vspace{-0.1in}
\subsection{Decoders: Identifying Valid Positions and Generating Concept Names}
\label{sec:decoder}

Given a masked position $v$, we now have the fused representations of its relational contexts $\mathbf{v}_{\text{fuse}}$ from the above encoders. We perform two tasks jointly to complete the taxonomy: one is to identify whether the candidate position is valid or not; the other is to generate the name of concept for the masked position.

\subsubsection{Task 1: Candidate Position Classification}

Given a candidate position, this task predicts whether it is valid, i.e., a concept is needed. If the position has a masked concept in the existing taxonomy, it is a valid position; otherwise, this invalid position is produced by negative sampling. We use a three-layer feed forward neural network (FFNN) to estimate the probability of being a valid position with the fused representations: $p_{\text{valid}}(v) = \textsc{FFNN}(\mathbf{v}_{\text{fuse}})$. The loss term on the particular position $v$ is based on cross entropy:
\begin{equation}
    L_1(\Theta; v) = -(y_v \cdot \log(p_{\text{valid}}(v)) + (1-y_v) \cdot \log(1-p_{\text{valid}}(v))), \nonumber
\end{equation}
where $y_v = 1$ when $v$ is valid as observed; otherwise, $y_v = 0$.

\paragraph{Negative sampling:} Suppose a valid position is sampled by masking an existing concept $v \in \mathcal{V}_0$, whose set of parent/ancestor nodes is denoted by $\mathcal{V}_p (v)$ and set of child/descendant nodes is denoted by $\mathcal{V}_c (v)$. We create $r_{\text{neg}}$ negative samples (i.e., invalid positions) by replacing one concept in either $\mathcal{V}_p (v)$ or $\mathcal{V}_c (v)$ by a random concept in $\mathcal{V}_{\text{neg}}(v) = \mathcal{V}_0 \setminus (\mathcal{V}_p (v) \cup \mathcal{V}_c (v) \cup \{v\})$. We will investigate the effect of negative sampling ratio $r_{\text{neg}}$ in experiments.

\subsubsection{Task 2: Concept Name Generation}

We use a Gated Recurrent Unit (GRU)-based decoder to generate the name of concept token by token which is a variable-length sequence $v = (v_1, v_2, \dots, v_{T_v})$.
As shown at the right of Figure~\ref{fig:framework}, we add a special token [EOS] to  indicate the concept generation is finished.

We initialize the hidden state of the decoder as $\mathbf{v}_0 = \mathbf{h}_{\text{seq}}$. Then the conditional language model works as below:
\begin{eqnarray}
    & & \mathbf{v}_t = \textsc{GRU}(v_{t-1}, \mathbf{v}_{t-1} \oplus \mathbf{v}_{\text{graph}}), \nonumber \\
    & & p(v_t | v_{t-1}, \dots, v_1, \mathbf{v}_{\text{fuse}}) = \textsc{Readout}(v_{t-1}, \mathbf{v}_t, \mathbf{v}_{\text{fuse}}), \nonumber
\end{eqnarray}
where $\textsc{Readout}(\cdot)$ is a nonlinear multi-layer function that predicts the probability of token $v_t$. Then this task can be regarded as a sequential multi-class classification problem. The loss term is:
\begin{equation}
    L_2(\Theta; v) = -\log(p(v | \mathbf{v}_{\text{fuse}})) = -\sum^{T_v}_{t=1} \log(p(v_t | v_{<t}, \mathbf{v}_{\text{fuse}})), \nonumber
\end{equation}
where $v$ is the target sequence (i.e., the concept name) and $\mathbf{v}_{\text{fuse}}$ is the fused relational representations of the masked position.

\paragraph{Pre-training:} To perform well in Task 2, a model needs the ability of predicting tokens in a concept's name. So, we pre-train the model with the task of predicting masked concept's tokens (MCT) in sentences of text corpus $\mathcal{D}$.
We find all the sentences in $\mathcal{D}$ that contain at least one concept in the existing taxonomy $\mathcal{T}_0$.
Given a sentence $X = (x_1, x_2, \dots, x_n)$ where $(x_s, \dots, x_e) = v = (v_1, \dots, v_{T_v})$ is an existing concept. Here a token $x_m$ is masked ($s \leq m \leq e$) and predicted using all past and future tokens. The loss function is:
\begin{equation}
    L_{\text{MCT}}(\Theta; x_m) = -\log(p(x_m | x_1, \dots, x_{m-1}, x_{m+1}, \dots, x_n)). \nonumber
\end{equation}

\subsubsection{Joint Training}

The joint loss function is a weighted sum of the loss terms of the above two tasks:
\begin{equation}
    L = \sum_{v \in \mathcal{V}_0} \left( L_1(\Theta; v) + \lambda L_2(\Theta; v) \right) + \sum^{r_{\text{neg}}}_{i = 1} \mathbb{E}_{v_{\text{neg}} \sim \mathcal{V}_{\text{neg}}(v)} L_1(\Theta; v_{\text{neg}}), \nonumber
\end{equation}
where $\lambda$ is introduced as a hyper-parameter to control the importance of Task 2 concept name generation.

\subsection{\methodplus: Enhancing Extraction-based Methods with \method}
\label{sec:methodology2}

While \method is designed to replace the process of extracting new concepts by concept generation, \methodplus is an alternative solution when the set of new concepts is given and of high quality.
\methodplus can use any extraction-based method \cite{shen2020taxoexpan,shang2020nettaxo,shang2020taxonomy,zhang2018taxogen,zhang2021taxonomy,manzoor2020expanding,yu2020steam,wang2021enquire} as the main framework and iteratively expand the set of new concepts using concept generation (i.e., \method) to continuously improve the taxonomy completeness. We choose \textsc{TaxoExpan} as the extraction-based method \methodplus \cite{shen2020taxoexpan}.

The details of \methodplus are as follows. We start with an existing taxonomy $T_0 = (\mathcal{V}_0, \mathcal{E}_0)$ and a \emph{given} set of new concepts $C_0$. During the $i$-th iteration, we first use \method to \emph{generate} a set of new concepts, and then expand the set of new concepts and use the extraction-based method to update the taxonomy ($i \geq 1$):
\vspace{-0.05in}
\begin{eqnarray}
    {T'}_{i-1} = (\mathcal{V'}_{i-1}, \mathcal{E'}_{i-1}) & \leftarrow & \method(T_{i-1}), \nonumber \\
    C_i & = & (C_{i-1} \cup \mathcal{V'}_{i-1}) \setminus \mathcal{V}_{i-1}, \nonumber \\
    T_i & \leftarrow & \textsc{ExtractionMethod}(T_{i-1}, C_i). \nonumber
\end{eqnarray}
The iterative procedure terminates when $C_i == \emptyset$. In this procedure, we have two hyperparameters:
\begin{enumerate}
\item Concept quality threshold $\tau$: \method predicts the probability of being a valid position $p_{\text{valid}}(v)$ which can be considered as the quality of the generated concept $v$. We have a constraint on adding generated concepts to the set: $p_{\text{valid}}(v) \geq \tau$, for any $v \in C_i$. When $\tau$ is bigger, the process is more cautious: fewer concepts are added each iteration.
\item Maximum number of iterations ${max}_{\text{iter}}$: An earlier stop is more cautious but may cause the issue of low recall.
\end{enumerate}
% We investigate the effect of these hyperparameters in experiments.

%% file: 5ExperimentDesign.tex
\vspace{-0.1in}
\section{Experimental Settings}
\label{sec:5ExperimentDesign}

In this work, we propose \method and \methodplus to complete taxonomies through concept generation.
We conduct experiments to answer the following research questions (RQs):
\begin{itemize}
    \item \textbf{RQ1:} Do the proposed approaches perform better than existing methods on \emph{taxonomy completion}?
    \item \textbf{RQ2:} Given \emph{valid} positions in an existing taxonomy and a corresponding large text corpus, which method \emph{produce more accurate concept names}, the proposed concept generation or existing extraction-and-filling methods?
    \item \textbf{RQ3:} How do the components and hyperparametersa impact the performance of \method?
\end{itemize}

\begin{table}[t]
    \centering
    \Scale[0.95]{\begin{tabular}{|l|r|r|r|r|}
    \hline
    & \#Concepts & \#Tokens & \#Edges & Depth \\
    \hline
    \rowcolor{gray!12} \multicolumn{5}{|c|}{Computer Science domains (Corpus: DBLP)} \\
    \hline
    MAG-CS & 29,484 & 16,398 & 46,144 & 6 \\
    {\small (found in corpus)} & 18,338 (62.2\%) & 13,914 (84.9\%) & & \\ \hline
    OSConcepts & 984 & 967 & 1,041 & 4 \\ \hline
    DroneTaxo & 263 & 247 & 528 & 4 \\
    \hline
    \rowcolor{gray!12} \multicolumn{5}{|c|}{Biology/Biomedicine domains: (Corpus: PubMed)} \\
    \hline
    MeSH & 29,758 & 22,367 & 40,186 & 15 \\
    {\small (found in corpus)} & 14,164 (47.6\%) & 22,193 (99.2\%) & & \\ \hline
    SemEval-Sci & 429 & 573 & 452 & 8 \\ \hline
    SemEval-Env & 261 & 317 & 261 & 6 \\
    \hline
    \end{tabular}}
    \caption{Statistics of six taxonomy data sets.}
    \label{tab:datasets}
    \vspace{-0.3in}
\end{table}

\vspace{-0.1in}
\subsection{Datasets}

Table~\ref{tab:datasets} presents the statistics of six taxonomies from two different domains we use to evaluate the taxonomy completion methods:
\begin{itemize}
\item \textbf{MAG-CS \cite{sinha2015overview}:} The Microsoft Academic Graph (MAG) taxonomy has over 660 thousand scientific concepts and more than 700 thousand taxonomic relations. We follow the processing in \textsc{TaxoExpan} \cite{shen2020taxoexpan} to select a partial taxonomy under the ``computer science'' (CS) node.
\item \textbf{OSConcepts \cite{peterson1985operating}:} It is a taxonomy manually crated in a popular textbook ``Operating System Concepts" for OS courses.
\item \textbf{DroneTaxo:}\footnote{\url{http://www.dronetology.net/}} is a human-curated hierarchical ontology on unmanned aerial vehicle (UAV).
\item \textbf{MeSH \cite{lipscomb2000medical}:} It is a taxonomy of medical and biological terms suggested by the National Library of Medicine (NLM).
\item \textbf{SemEval-Sci} and \textbf{SemEval-Env \cite{bordea2016semeval}:} They are from the shared task of taxonomy construction in SemEval2016. We select two scientific-domain taxonomies (``science'' and ``environment''). Both datasets have been used in \cite{yu2020steam,wang2021enquire}.
\end{itemize}

We use two different corpora for the two different domains of data:
(1) \textbf{DBLP corpus} has about 156K paper abstracts from the computer science bibliography website;
(2) \textbf{PubMed corpus} has around 30M abstracts on MEDLINE.
We observe that on the two largest taxonomies, around a half of concept names and a much higher percentage of unique tokens can found \emph{at least once} in the corpus, which indicates a chance of generating rare concept names token by token. The smaller taxonomies show similar patterns.

\input{Tables/baseline_exp1}

\vspace{-0.1in}
\subsection{Evaluation Methods}

We randomly divide the set of concepts of each taxonomy into training, validation, and test sets at a ratio of 3:1:1. We build ``existing'' taxonomies with the training sets following the same method in \cite{zhang2021taxonomy}.  
To answer RQ1, we use Precision, Recall, and F1 score to evaluate the completeness of taxonomy. The Precision is calculated by dividing the number of correctly inserted concepts by the number of total inserted concepts, and Recall is calculated by dividing the the number of correct inserted concepts by the number of total concepts.
For RQ2, we use accuracy (Acc) to evaluate the quality of generated concepts. For IE models, we evaluate what percent of concepts in taxonomy can be correctly extracted/generated when a position is given. We use Uni-grams (Acc-Uni), and Accuracy on Multi-grams (Acc-Multi) for scenarios where dataset contains only Uni-grams and multi-gram concepts.

\input{Tables/baseline_exp3_PRF}

\vspace{-0.1in}
\subsection{Baselines}

This work proposes the first method that generates concepts for taxonomy completion.
Therefore, 
We compare \method and \methodplus with state-of-the-art extraction-based methods below:
\begin{itemize}
\item \textbf{\textsc{HiExpan} \cite{shen2018hiexpan}} uses textual patterns and distributional similarities to capture the ``isA'' relations and then organize the extracted concept pairs into a DAG as the output taxonomy.
\item \textbf{\textsc{TaxoExpan} \cite{shen2020taxoexpan}} adopts GNNs to encode the positional information and uses a linear layer to identify whether the candidate concept is the parent of the query concept.
\item \textbf{\textsc{Graph2Taxo} \cite{shang2020taxonomy}} uses cross-domain graph structure and constraint-based DAG learning for taxonomy construction.
\item \textbf{\textsc{STEAM} \cite{yu2020steam}} models the mini-path information to capture global structure information to expand the taxonomy.
\item \textbf{\textsc{ARBORIST} \cite{manzoor2020expanding}} is the state-of-the-art method for taxonomy expansion. It aims for taxonomies with heterogeneous edge semantics and adopts a large-margin ranking loss to guaranteed an upper-bound on the shortest-path distance between the predicted parents and actual parents.
\item \textbf{\textsc{TMN} \cite{zhang2021taxonomy}} is the state-of-the-art method for taxonomy completion. It uses a triplet matching network to match a query concept with (hypernym, hyponym)-pairs.
\end{itemize}
For RQ1, the extraction-based baselines as well as \methodplus are offered the concepts from the test set if they can be extracted from the text corpus but \emph{NOT} to the pure generation-based \method. 
For RQ2, given a \emph{valid} position in an existing taxonomy, it is considered as accurate if a baseline can extract the desired concept from text corpus and assign it to that position or if \method can generate the concept's name correctly.
% Hyperparameter settings are in Appendix.

%% file: Tables/baseline_exp1.tex
\begin{table}[t]
\begin{tabular}{|l|rrr|}
\hline
\cellcolor{gray!12} MeSH: & \multicolumn{1}{c}{\textbf{Acc}} & \multicolumn{1}{c}{Acc-Uni} & \multicolumn{1}{c|}{Acc-Multi} \\
\hline
\textsc{HiExpan} \cite{shen2018hiexpan} & 9.89 & 18.28 & 8.49 \\
\textsc{TaxoExpan} \cite{shen2020taxoexpan} & 16.32 & \underline{28.35} & 14.31 \\
\textsc{Graph2Taxo} \cite{shang2020taxonomy} & 10.35 & 19.02 & 8.90 \\
\textsc{STEAM} \cite{yu2020steam} & \underline{17.04} & 27.61 & \underline{15.26} \\
\textsc{ARBORIST} \cite{manzoor2020expanding} & 16.01 & 27.91 & 14.19 \\
\textsc{TMN} \cite{zhang2021taxonomy} & 16.53 & 27.52 & 14.85 \\
\hline
\textbf{\method} ($r_\text{neg} = 0$) & 26.72 & 28.13 & \textbf{26.29} \\
\textbf{\method} ($r_\text{neg} = 0.15$) & \textbf{26.93} & \textbf{31.34} & 26.07 \\
\hline
\end{tabular}
\caption{Evaluating the quality of generated concepts (RQ2).}
\label{tab:baseline_exp1}
\vspace{-0.3in}
\end{table}

%% file: Tables/baseline_exp3_PRF.tex
\begin{table}[t]
\begin{tabular}{|l|rrr|rrr|}
\hline
\cellcolor{gray!12} Largest & \multicolumn{3}{c|}{\textbf{MAG-CS}} & \multicolumn{3}{c|}{\textbf{MeSH}} \\
\cellcolor{gray!12} two: & \multicolumn{1}{c}{P} & \multicolumn{1}{c}{R} & \multicolumn{1}{c|}{\textbf{F1}} & \multicolumn{1}{c}{P} & \multicolumn{1}{c}{R} & \multicolumn{1}{c|}{\textbf{F1}} \\
\hline
\textsc{HiExpan} & 19.61 & 8.23 & 11.59 & 17.77 & 5.66 & 8.59 \\
\textsc{TaxoExpan} & 36.19 & 20.20 & 25.92 & \underline{\textbf{26.87}} & \underline{11.79} & \underline{16.39} \\
\textsc{Graph2Taxo} & 23.43 & 12.97 & 16.70 & 26.13 & 10.35 & 14.83 \\
\textsc{STEAM} & \underline{\textbf{36.73}} & \underline{23.42} & \underline{28.60} & 26.05 & 11.23 & 15.69 \\
\textsc{ARBORIST} & 29.72 & 15.90 & 20.72 & 26.19 & 10.76 & 15.25 \\
\textsc{TMN} & 28.82 & 23.09 & 25.64 & 23.73 & 9.84 & 13.91 \\
\hline
\textbf{\method} & 36.15 & 28.19 & 31.67 & 21.47 & 17.10 & 19.03  \\
\textbf{\methodplus} & 36.24 & \textbf{28.68} & \textbf{32.01} & 22.61 & \textbf{17.66} & \textbf{19.83} \\
\hline \hline
\cellcolor{gray!12} Computer & \multicolumn{3}{c|}{\textbf{OSConcepts}} & \multicolumn{3}{c|}{\textbf{DroneTaxo}} \\
\cellcolor{gray!12} Science: & \multicolumn{1}{c}{P} & \multicolumn{1}{c}{R} & \multicolumn{1}{c|}{\textbf{F1}} & \multicolumn{1}{c}{P} & \multicolumn{1}{c}{R} & \multicolumn{1}{c|}{\textbf{F1}} \\
\hline
\textsc{HiExpan} & 21.77 & 13.71 & 16.82 & 43.24 & 30.77 & 35.95 \\
\textsc{TaxoExpan} & 30.43 & \underline{21.32} & \underline{25.07} & \underline{60.98} & 48.08 & 53.77 \\
\textsc{Graph2Taxo} & 22.88 & 13.71 & 17.15 & 44.90 & 23.31 & 30.69 \\
\textsc{STEAM} & 30.71 & 19.79 & 24.07 & 58.33 & \underline{53.85} & \underline{56.00} \\
\textsc{ARBORIST} & \underline{\textbf{31.09}} & 18.78 & 23.42 & 52.38 & 42.31 & 46.81 \\
\textsc{TMN} & 30.65 & 19.29 & 23.68 & 47.72 & 40.38 & 43.74 \\
\hline
\textbf{\method} & 18.32 & 12.18 & 14.63 & 11.63 & 9.62 & 10.53 \\
\textbf{\methodplus} & 30.18 & \textbf{25.89} & \textbf{27.87} & \textbf{65.96} & \textbf{59.62} & \textbf{62.63} \\
\hline \hline
\cellcolor{gray!12} Biology/ & \multicolumn{3}{c|}{\textbf{SemEval-Sci}} & \multicolumn{3}{c|}{\textbf{SemEval-Env}} \\
\cellcolor{gray!12} Biomedicine: & \multicolumn{1}{c}{P} & \multicolumn{1}{c}{R} & \multicolumn{1}{c|}{\textbf{F1}} & \multicolumn{1}{c}{P} & \multicolumn{1}{c}{R} & \multicolumn{1}{c|}{\textbf{F1}} \\
\hline
\textsc{HiExpan} & 14.63 & 10.34 & 12.12 & 15.79 & 8.11 & 10.72 \\
\textsc{TaxoExpan} & 24.14 & \underline{29.17} & 26.42 & 23.07 & 16.22 & 19.05 \\
\textsc{Graph2Taxo} & 26.19 & 18.96 & 21.99 & 21.05 & 10.81 & 14.28 \\
\textsc{STEAM} & 35.56 & 27.58 & \underline{31.07} & \underline{46.43} & \underline{35.13} & \underline{39.99} \\
\textsc{ARBORIST} & \underline{\textbf{41.93}} & 22.41 & 29.21 & 46.15 & 32.43 & 38.09 \\
\textsc{TMN} & 34.15 & 24.14 & 28.29 & 37.93 & 29.73 & 33.33 \\
\hline
\textbf{\method} & 11.43 & 6.90 & 8.61 & 16.13 & 13.51 & 14.70 \\
\textbf{\methodplus} & 38.78 & \textbf{32.76} & \textbf{35.52} & \textbf{48.28} & \textbf{37.84} & \textbf{42.42} \\
\hline
\end{tabular}
\caption{Performance on taxonomy completion: Bold for the highest among all. Underlined for the best baseline. (RQ1)}
\label{tab:baseline_exp_PRF}
\vspace{-0.3in}
\end{table}

%% file: 6ExperimentResults.tex
\section{Experimental Results}
\label{sec:experiment}

% We report experimental results to answer the three RQs in Section~\ref{sec:5ExperimentDesign}.

\begin{figure}[t]
    \centering
    \includegraphics[width=0.9\linewidth]{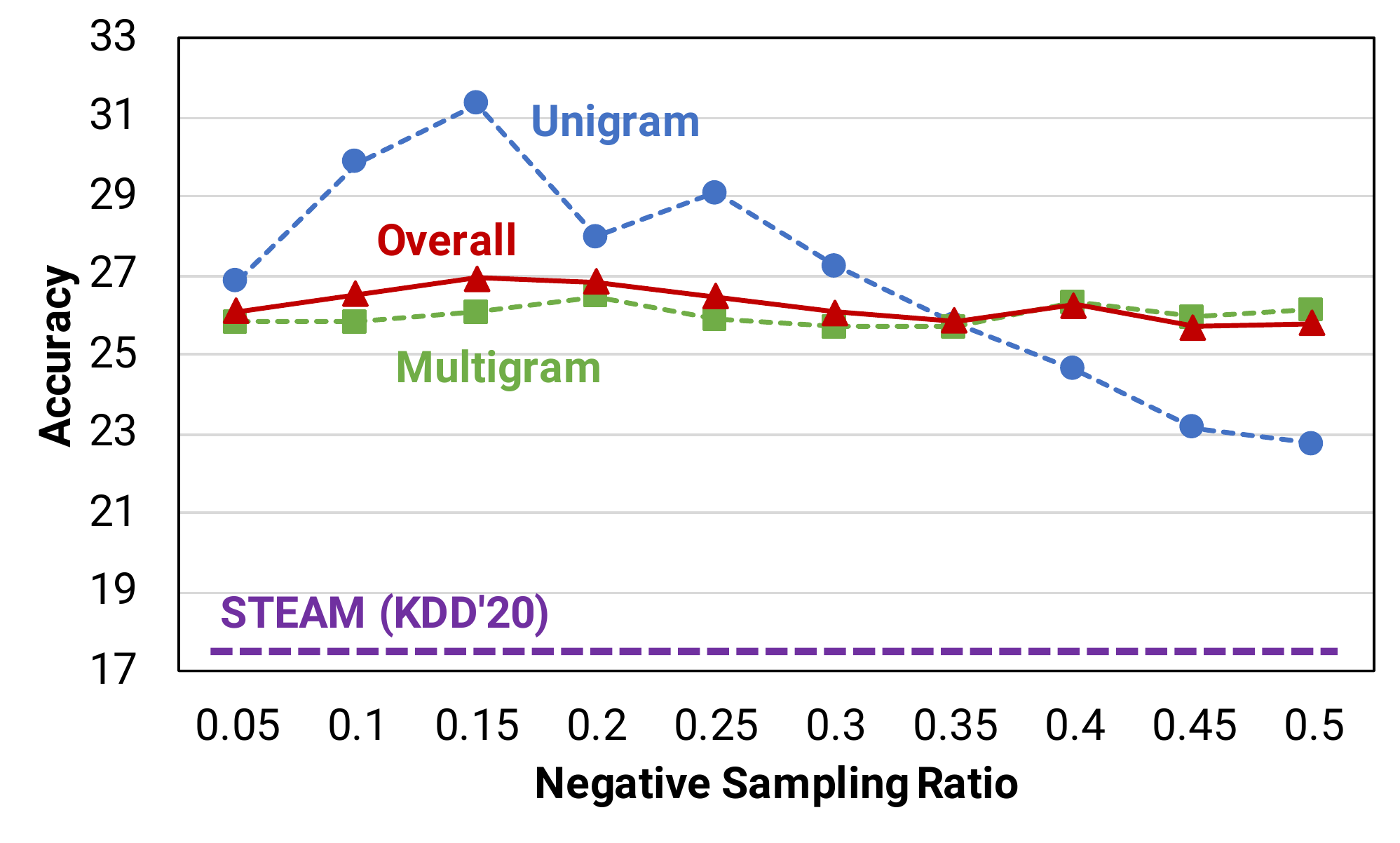}
    \vspace{-0.2in}
    \caption{Effect of negative sampling ratio $r_\text{neg}$ on the quality of concept name generation.}
    \label{fig:negative_sampling}
    \vspace{-0.2in}
\end{figure}

\subsection{RQ1: Taxonomy Completion}
\label{sub_sec:exp1}

% \begin{figure}[t]
%     \centering
%     \includegraphics[width=\linewidth]{Figures/results_neg.pdf}
%     \vspace{-0.35in}
%     \caption{Effect of negative sampling ratio $r_\text{neg}$ on the quality of concept name generation.}
%     \label{fig:negative_sampling}
%     \vspace{-17pt}
% \end{figure}

Table~\ref{tab:baseline_exp_PRF} presents the results on taxonomy completion.
We have three main observations. First, \textsc{TaxoExpan} and \textsc{STEAM} are either the best or the second best among all the baselines on the six datasets.
When \textsc{TaxoExpan} is better, the gap between the two methods in terms of F1 score is no bigger than 0.7\% (15.69 vs 16.39 on MeSH); when \textsc{STEAM} is better, its F1 score is at least 2.2\% higher than \textsc{TaxoExpan} (e.g., 28.60 vs 25.92 on MAG-CS). So, \textsc{STEAM} is generally a stronger baseline. This is because the sequential model that encodes the mini-paths from the root node to the leaf node learns useful information. The sequence encoders in our \method learn such information from the template-based sentences. \textsc{STEAM} loses to \textsc{TaxoExpan} on MeSH and OSConcepts due to the over-smoothing issue of too long mini-paths.
We also find that \textsc{ARBORIST} and \textsc{TMN} do not perform better than \textsc{STEAM}. This indicates that GNN-based encoder in \textsc{STEAM} captures more structural information (e.g., sibling relations) than \textsc{ARBORIST}'s shortest-path distance and \textsc{TMN}'s scoring function based on hypernym-hyponym pairs.

Second, on the largest two taxonomies MAG-CS and MeSH, \method outperforms the best extraction-based methods \textsc{STEAM} and \textsc{TaxoExpan} in terms of recall and F1 score. This is because with the module of concept generation, \method can produce more desired concepts beyond those that are extracted from the corpus.
Moreover, \method uses the fused representations of relational contexts.
Compared with \textsc{STEAM}, \method encodes both top-down and bottom-up subgraphs.
When \textsc{TaxoExpan} considers one-hop ego-nets of a query concept, \method aggregates multi-hop information into the fused hidden states.

Third, on the smaller taxonomies \method performs extremely bad due to the insufficient amount of training data (i.e., fewer than 600 concepts). Note that we assumed the set of new concepts was given for all the extraction-based methods, while \method was not given it and had to rely on name generation to ``create'' the set. In a fair setting -- when we allow \method to use the set of new concepts, then we have -- \methodplus performs consistently better than all the baselines. This is because it takes advantages of both the concept generation and extraction-and-fill methodologies.
% \begin{figure}[t]
%     \centering
%     \includegraphics[width=\linewidth]{Figures/results_neg.pdf}
%     \vspace{-0.35in}
%     \caption{Effect of negative sampling ratio $r_\text{neg}$ on the quality of concept name generation.}
%     \label{fig:negative_sampling}
%     \vspace{-17pt}
% \end{figure}

\input{Tables/seq_encoder}

\subsection{RQ2: Concept Name Generation}

Table~\ref{tab:baseline_exp1} presents the results on concept name generation/extraction: Given a \emph{valid} position on an existing taxonomy, evaluate the accuracy of the concept names that are (1) extracted and filled by baselines or (2) generated by \method. Our observations are:

First, among all baselines, \textsc{STEAM} achieves the highest on multi-gram concepts, and \textsc{TaxoExpan} achieves the highest accuracy on uni-gram concepts. This is because the mini-path from root to leaf may encode the naming system for a multi-gram leaf node.

Second, the accuracy scores of \textsc{TaxoExpan}, \textsc{STEAM}, \textsc{ARBORIST}, and \textsc{TMN} are not significantly different from each other (within 1.03\%). This indicates that these extraction-based methods are unfortunately limited by the presence of concepts in corpus.

Third, compared \method (the last two lines in the table) with \textsc{STEAM}, we find \method achieves 9.7\% higher accuracy. This is because \method can assemble frequent tokens into infrequent or even unseen concepts.

Then, \emph{do negative samples help learn to generate concept names?} In Figure~\ref{fig:negative_sampling}, we show the accuracy of \method given different values of negative sampling ratio $r_\text{neg}$.

First, We observe that \method performs consistently better than the strongest baseline \textsc{STEAM} in this case. And the overall accuracy achieves the highest when $r_\text{neg} = 0.15$. From our point of view, the negative samples accelerate the convergence at early stage by providing a better gradient descending direction for loss function. However, too many negative samples would weaken the signal from positive examples, making it difficult for the model to learn knowledge from them.

Second, we find that the uni-gram and multi-gram concepts have different kinds of sensitivity to the ratio but comply to the same trend. Generally uni-grams have higher accuracy because generating fewer tokens is naturally an easier task; however, they take a smaller percentage of the data. So the overall performance is closer to that on multi-grams. And our \method focuses on generating new multi-gram concepts.

\input{Tables/graph_encoder}

\subsection{RQ3: Ablation Studies}

In this section, we perform ablation studies to investigate two important designs of \method: (3.1) Sequence encoders for sentence-based relational contexts; (3.2) Graph encoders for subgraph-based contexts. We use MeSH in these studies, so based on Table~\ref{tab:baseline_exp_PRF}, the F1 score of \method is \emph{19.03}.

% In this section, we perform ablation studies to investigate three important designs of \method: (3.1) Sequence encoders for sentence-based relational contexts; (3.2) Graph encoders for subgraph-based contexts; (3.3) Representation fusion methods. We use MeSH in these studies, so based on Table~\ref{tab:baseline_exp_PRF}, the F1 score of \method is \emph{19.03}.

\subsubsection{Sequence encoders for sentence-based relational contexts}

We consider three types of encoders: GRU, Transformer, and SciBERT,  pre-trained by a general masked language model (MLM) task on massive scientific corpora.
We use our proposed task (Masked Concept's Tokens) to pre-train GRU/Transformer and fine-tune SciBERT.
We add sentences that describe 1-hop relations (i.e., parent or child), sibling relations, 2-hops relations, and 3-hops relations step by step.
Table~\ref{tab:seq-encoder} presents the results of all the combinations. Our observations are as follows.

First, the pre-training process on related corpus is useful for generating a concept's name at a candidate position in an existing taxonomy. The pre-training task is predicting a token in an existing concept, strongly relevant with the target task. We find that pre-training improves the performance of the three sequence encoders.

Second, surprisingly we find that GRU performs \emph{slightly better} than Transformer and SciBERT (19.03 vs 18.53 and 18.87). The reason may be that the sentence templates that represent relational contexts of a masked position always place two concepts in a relation at the beginning or end of the sentence. Because GRU encodes the sentences from both left-to-right and right-to-left directions, it probably represents the contexts better than the attention mechanisms in Transformer and SciBERT. SciBERT is pre-trained on MLM and performs better than Transformer.

Third, it is not having all relations in 3-hops neighborhood of the masked concept that generates the highest F1 score. On all the three types of sequence encoders, we find that the best collection of constructed sentences are those that describe 1-hop relations (i.e., parent and child) and sibling relations which is a typical relationship of two hops. Because 1-hop ancestor (parent), 2-hop ancestor (grandparent), and 3-hop ancestor are all the ``class'' of the masked concept, if sentences were created for all these relations, sequence encoders could not distinguish the levels of specificity of the concepts. Similarly, it is good to represent only the closest type of descendant relationships (i.e., child) as ``subclass'' of the masked concept. And sibling relations are very useful for generating concept names. For example, in Figure~\ref{fig:idea}, ``nucleic acid denaturation'' and ``nucleic acid renaturation'' have similar naming patterns when they are sibling concepts.

\input{Tables/neg_sense}

\subsubsection{Graph encoders for subgraph-based relational contexts}

We consider four types of GNN-based encoders: GAT \cite{velivckovic2018graph}, GCN \cite{kipf2016semi}, GraphSAGE \cite{hamilton2017inductive}, and GraphTransformer \cite{koncel2019text}.
We add 1-hop, 2-hop, and 3-hop relations step by step in constructing the top-down and bottom-up subgraphs.
The construction forms either undirected or directed subgraphs in the information aggregation GNN algorithms.
Table~\ref{tab:graph-encoder} presents the results. Our observations are as follows.

First, we find that encoding directed subgraphs can achieved a better performance than encoding undirected subgraphs for all the four types of graph encoders. This is because the directed subgraph can represent asymmetric relations. For example, it can distinguish parent-child and child-parent relations. In directed subgraphs, the edges always point from parent to child while such information is missing in undirected graphs.

Second, the best graph encoder is GraphTransformer and the second best is Graph Attention Network (GAT). They both have the attention mechanism which plays a significant role in aggregating information from top-down and bottom-up subgraphs for generating the name of concept. GraphTransformer adopts the Transformer architecture (of \emph{all} attention mechanism) that can capture the contextual information better and show stronger ability of generalization than GAT.

Third, we find that all the types of graph encoders perform the best with 2-hops subgraphs. The reason may be that the GNN-based architectures cannot effectively aggregate multi-hop information. In other words, they suffer from the issue of over smoothing when they use to encode information from 3-hops neighbors.

%% file: Tables/seq_encoder.tex
\begin{table}[t]
\begin{tabular}{|l|l|c|cc|c|}
\hline
\cellcolor{gray!12} MeSH: & PT & 1 hop & \multicolumn{2}{c|}{+ 2 hops} & + 3 hops \\
\cellcolor{gray!12} (F1 score) & & & Sibling & Grand-p/c & \\
\hline
GRU & $\times$ & 18.19 & 18.29 & 18.92 & 17.41 \\
& $\checkmark$ & 18.35 & \textbf{19.03} & 18.40 & 17.49 \\
\hline
Transformer & $\times$ & 17.89 & 18.13 & 17.97 & 17.04 \\
\cite{vaswani2017attention} & $\checkmark$ & 18.02 & 18.53 & 18.19 & 17.07 \\
\hline
SciBERT & $\times$ & 18.05 & 18.16 & 18.12 & 17.29 \\
\cite{beltagy2019scibert} & $\checkmark$ & 18.11 & 18.87 & 18.23 & 17.41 \\
\hline
\end{tabular}
\caption{Ablation study on sequence encoders for sentence-based relational contexts. (PT is for Pre-training. Grand-p/c is for grandparent/grandchild.)}
\label{tab:seq-encoder}
\vspace{-0.3in}
\end{table}

% \begin{table*}[t]
% \caption{Comparison of three sequence encoders (GRU, SciBERT, Transformer) for encoding synthetic sentences. We investigated which kinds of relationships are most suitable to be expressed as synthetic sentences for sentence-based context encoder. The 2-hop relations include grandparent-grandchild and sibling relations. The results shows that GRU with 1-hop relation and 2-hop relation \emph{without} grandparent-grandchild relation is the best selection. }
% \begin{tabular}{l|cccc|cccc|cccc}
% \toprule
%  & \multicolumn{4}{c|}{\textbf{F1}} & \multicolumn{4}{c|}{\textbf{F1-Uni}} & \multicolumn{4}{c}{\textbf{F1-Multi}} \\ \midrule
%  & \textbf{1-hop} & \textbf{\begin{tabular}[c]{@{}c@{}}2-hop \\ (sib. only)\end{tabular}} & \textbf{2-hop} & \textbf{3-hop} & \textbf{1-hop} & \textbf{\begin{tabular}[c]{@{}c@{}}2-hop \\ (sib. only)\end{tabular}} & \textbf{2-hop} & \textbf{3-hop} & \textbf{1-hop} & \textbf{\begin{tabular}[c]{@{}c@{}}2-hop \\ (sib. only)\end{tabular}} & \textbf{2-hop} & \textbf{3-hop} \\ \midrule
% \textbf{GRU} & 16.83 & \underline{\textbf{18.78}} & 17.45 & 15.39 & 16.26 & \underline{\textbf{19.65}} & 19.62 & 14.12 & 16.82 & \underline{\textbf{18.64}} & 17.12 & 15.59 \\
% \textbf{SciBERT} & 16.55 & 18.51 & 17.34 & 15.21 & 16.37 & 19.10 & 18.79 & 14.95 & 16.92 & 18.43 & 16.57 & 15.52 \\
% \textbf{Transformer} & 16.40 & 18.15 & 16.92 & 15.04 & 15.96 & 18.93 & 18.56 & 14.79 & 16.42 & 18.16 & 16.21 & 15.09 \\ \bottomrule
% \end{tabular}
% \label{tab:seq-encoder}
% \end{table*}

%% file: Tables/graph_encoder.tex
\begin{table}[t]
\label{tab:graph-encoder}
\centering
\begin{tabular}{|l|l|ccc|}
\hline
\cellcolor{gray!12} MeSH: (F1 score) & U/D & 1 hop & + 2 hops & + 3 hops \\
\hline
GAT \cite{velivckovic2018graph} & U & 18.17 & 18.43 & 17.11 \\
& D & 18.29 & 18.94 & 17.39 \\
\hline
GCN \cite{kipf2016semi} & U & 18.10 & 18.27 & 17.09 \\
& D & 18.21 & 18.35 & 17.15 \\
\hline
GraphSAGE \cite{hamilton2017inductive} & U & 18.12 & 18.36 & 17.05 \\
& D & 18.19 & 18.66 & 17.20 \\
\hline
GraphTransformer \cite{koncel2019text} & U & 18.22 & 18.79 & 17.13 \\
 & D & 18.35 & \textbf{19.03} & 17.49 \\
\hline
\end{tabular}
\caption{Ablation study on graph encoders for subgraph contexts on the taxonomy. (U/D is for undirected/directed subgraph settings in GNN-based aggregation.)}
\label{tab:graph-encoder}
\vspace{-0.35in}
\end{table}

%% file: Tables/neg_sense.tex
\begin{table}[t]
\begin{tabular}{|c|c|l|rrr|}
\hline
Sentence- & Graph- & Fusion method & \multirow{2}{*}{P} & \multirow{2}{*}{R} & \multirow{2}{*}{F1} \\
based & based & $\textsc{Fuse}(\cdot)$ & & & \\
\hline
$\checkmark$ & $\times$ & - & 23.62 & 14.83 & 18.22 \\
\hline
$\times$ & $\checkmark$ & - & 7.14 & 9.06 & 7.98 \\
\hline
\multirow{4}{*}{$\checkmark$} & \multirow{4}{*}{$\checkmark$} & {Mean Pooling} & 20.29 & 17.10 & 18.56 \\
\cline{3-6}
& & {Max Pooling} & 18.36 & 17.25 & 17.79 \\
\cline{3-6}
& & {Attention} & 20.75 & 17.20 & 18.81 \\ 
\cline{3-6}
& & {Concatenation} & 21.47 & 17.10 & \textbf{19.03} \\
\hline
\end{tabular}
\caption{Ablation study on the methods of fusing sentence-based and graph-based relational representations.}
\label{tab:neg-sense}
\vspace{-0.1in}
\end{table}

% \begin{table}[h]
% \caption{Different fusion strategies (graph information embedding \& sequence information embedding) for concept position classification module. We select the negative sampling ratio which makes \textsc{OntoGen} achieved the best performance. The experimental result shows that the best fusion strategy is concatenation. Also, all this strategy can get the better performance than IE-based models.}
% \begin{tabular}{lccc}
% \toprule
%  & \textbf{F1} & \textbf{F1-Uni} & \textbf{F1-Multi} \\  \midrule
% \textbf{Mean Pooling} & 18.56 & 17.40 & 18.63 \\
% \textbf{Max Pooling} & 17.79 & 17.13 & 18.25 \\
% \textbf{Attention} & 18.81 & 17.86 & \underline{\textbf{18.84}} \\ 
% \textbf{Concatenation} & \underline{\textbf{19.03}} & \underline{\textbf{22.14}} & 18.42 \\ \bottomrule
% \end{tabular}
% \label{tab:neg-sense}
% \end{table}

%% file: 2RelatedWork.tex
\section{Related Work}
\label{sec:relatedWork}

% Our work is closely related to the following three research topics.

\subsection{Taxonomy Construction}

Many methods used a two-step scheme: (1) extracted hypernym-hyponym pairs from corpus, then (2) organized the extracted relations into hierarchical structures.
Pattern-based \cite{nakashole2012patty,wu2012probase,zeng2019faceted} and embedding-based methods \cite{luu2016learning,jiang2019role} were widely used in the first step.
The second step was often considered as graph optimization and solved by maximum spanning tree \cite{bansal2014structured}, optimal branching \cite{velardi2013ontolearn},
and minimum-cost flow \cite{gupta2017taxonomy}.
Mao \emph{et al.} used reinforcement learning to organize the hypernym-hyponym pairs by optimizing a holistic tree metric as a reward function over the training taxonomies \cite{mao2018end}.
% Shang \emph{et al.} transferred knowledge from an existing taxonomy to generate one for a new domain \cite{shang2020taxonomy}.

\subsection{Taxonomy Expansion}

These methods aimed at collecting emergent terms and placing them at appropriate positions in an existing taxonomy. Aly \emph{et al.} adopted hyperbolic embedding to expand and refine an existing taxonomy \cite{aly2019every}. Shen \emph{et al.} \cite{shen2018hiexpan} and Vedula \emph{et al.} \cite{vedula2018enriching} applied semantic patterns to determine the position of the new terms. Fauceglia \emph{et al.} used a hybrid method to combine lexico-syntactic patterns, semantic web, and neural networks \cite{fauceglia2019automatic}. Manzoor \emph{et al.} proposed a joint-learning framework to simultaneously learn latent representations for concepts and semantic relations \cite{manzoor2020expanding}. Shen \emph{et al.} proposed a position-enhanced graph neural network to encode the relative position of each term \cite{shen2020taxoexpan}. Yu \emph{et al.} applied a mini-path-based classifier instead of hypernym attachment \cite{yu2020steam}.

\subsection{Keypharse Generation}

This is the most relevant task with the proposed concept name generation in taxonomies. Meng \emph{et al.} \cite{meng2017deep,meng2020empirical} applied Seq2Seq to generate keyphrases from scientific articles. Ahmad \emph{et al.} proposed a joint learning method to select, extract, and generate keyphrases \cite{ahmad2020select}.
Our approaches combine textual and taxonomic information to generate the concept names accurately.

%% file: 7Conclusions.tex
\section{Conclusions}
\label{sec:conclusions}

In this work, we proposed \method to enhance taxonomy completion by identifying the positions in existing taxonomies that need new concepts and generating the concept names. It learned position embeddings from both graph-based and language-based relational contexts. Experimental results demonstrated that \method improves the completeness of real-world taxonomies over extraction-based methods.

%% file: aAppendix.tex
\subsection{Sentence encoders}

The descriptions of three sentence encoders we use are as below.
\begin{enumerate}
\item BiGRU: Given a sentence $s \in \mathcal{S}(v)$, we denote $n$ as the length of $s$ which is also the position index of [MASK]. A three-layered Bidirectional Gated Recurrent Units (GRU) network creates the hidden state of the [MASK] token as $\mathbf{h}(s) = \overrightarrow{\mathbf{h}_n (s)} \oplus \overleftarrow{\mathbf{h}_n (s)}$, where $\oplus$ denotes vector concatenation.
\item Transformer: It is an encoder-decoder architecture with only attention mechanisms without any Recurrent Neural Networks (e.g., GRU). The attention mechanism looks at an input sequence and decides at each step which other parts of the sequence are important. The hidden state is that of the [MASK] token: $\mathbf{h}(s) = \mathbf{h}_n (s)$.
\item SciBERT: It is a bidirectional Transformer-based encoder \cite{beltagy2019scibert} that leverages unsupervised pretraining on a large corpus of scientific publications (from \url{semanticscholar.org}) to improve performance on scientific NLP tasks such as sequence tagging, sentence classification, and dependency parsing.
\end{enumerate}

\subsection{Baselines Implementation}

\textsc{HiExpan}, \textsc{TaxoExpan}, \textsc{STEAM}, and \textsc{ARBORIST} are used for taxonomy expansion. Given a subgraph which includes the ancestor, parent, and sibling nodes as the seed taxonomy, the final output is a taxonomy expanded by the target concepts. For taxonomy completion methods, we set the threshold for deciding whether to add the target node to the specified position of taxonomy as 0.5. For \textsc{Graph2Taxo}, we input the graph-based context within two hops from the target concept in order to construct the taxonomy structure. To evaluate \textsc{Graph2Taxo}, we check the parent and child nodes of the target concept. When both sets matched the ground truth, the prediction result is marked as correct.

\subsection{Hyperparameter Settings}

We use stochastic gradient descent (SGD) with momentum as our optimizer.
We applied a scheduler with ``warm restarts'' that reduces the learning rate from 0.25 to 0.05 over the course of 5 epochs as a cyclical regiment.
Models are trained for 50 epochs based on the validation loss.
All the dropout layers used in our model had a default ratio of 0.3.
The number of dimensions of hidden states is 200 for sequence encoders and 100 for graph encoders, respectively.
We search for the best loss weight $\lambda$ in \{0.1, 0.2, 0.5, 1, 2, 5\} and set it as 2.
We set $r_\text{neg} = 0.15$, $\tau = 0.8$, and $\max_{\text{iter}} = 2$ as default. Their effects will be investigated in the experiments.

% === for future use (journal?) === %
% \begin{figure}[t]
%     \centering
%     \includegraphics[width=\linewidth]{Figures/heatmap-f1_v4.pdf}
%     \caption{Effect of $\tau$ and ${max}_{\text{iter}}$ on the performance of \methodplus in terms of F1 score on MeSH. The GenTaxo++ result reported in Table~\ref{tab:baseline_exp_PRF} is highlighted as red.}
%     \label{fig:heatmap}
% \end{figure}

% \subsection{Hyperparameter Analysis in \textsc{GenTaxo++}}

% Figure~\ref{fig:heatmap} presents the F1 score of \methodplus under different values of quality threshold $\tau$ and maximum number of iterations ${max}_{\text{iter}}$. First, quality control in the iterative process is important. While the performance is not good when $\tau$ is too small or too big, choosing a proper value is not very sensitive. The range between 0.8 and 0.9 is good, and we choose the best $\tau$ at 0.8. Second, the best ${max}_{\text{iter}}$ should not be a big number. The main reason is error propagation -- as the number of iteration increases, the false positives from previous iterations will affect the next-step model training.
% === for future use (journal?) === %

% \section{Reproducibility}

% Code and data repository can be found at \url{https://github.com/QingkaiZeng/GenTaxo}.